\PassOptionsToPackage{table,xcdraw}{xcolor}
\documentclass[11pt]{article}
\usepackage[preprint]{acl}
\usepackage{times}
\usepackage{latexsym}
\usepackage[T1]{fontenc}
\usepackage[utf8]{inputenc}
\usepackage{microtype}
\usepackage{inconsolata}

\usepackage{amssymb}
\usepackage{amsmath}
\usepackage{booktabs}
\usepackage{multirow}
\usepackage{xcolor}
\usepackage{colortbl}
\usepackage{graphicx}
\usepackage{subcaption}

\title{Resource-Limited Joint Multimodal Sentiment Reasoning and Classification via Chain-of-Thought Enhancement and Distillation}

\author{
Haonan Shangguan\thanks{Equal contribution.} , 
Xiaocui Yang\footnotemark[1]\thanks{Corresponding author.}, 
Shi Feng, 
Daling Wang, \\
\textbf{Yifei Zhang, }
\textbf{Feiliang Ren, }
\textbf{Ge Yu} \\
School of Computer Science and Engineering, Northeastern University, Shenyang, China\\
\texttt{yangxiaocui@mail.neu.edu.cn}}

\hypersetup{
  pdftitle={Resource-Limited Joint Multimodal Sentiment Reasoning and Classification via Chain-of-Thought Enhancement and Distillation},
  pdfauthor={Haonan Shangguan, Xiaocui Yang, Shi Feng, Daling Wang, Yifei Zhang, Feiliang Ren, Ge Yu}
}

\begin{document}
\maketitle
\begin{abstract}
Current approaches for Multimodal Sentiment Analysis (MSA) primarily leverage the knowledge and reasoning capabilities of parameter-heavy (Multimodal) LLMs for classification, overlooking autonomous multimodal sentiment reasoning generation in resource-constrained environments.
In this paper, we focus on the Resource-Limited Joint Multimodal Sentiment Reasoning and Classification task, JMSRC, which simultaneously performs multimodal sentiment reasoning chain generation and sentiment classification only with a lightweight model.
We propose a Multimodal Chain-of-Thought Reasoning Distillation model, MulCoT-RD, designed for JMSRC that employs a "Teacher-Assistant-Student" distillation paradigm to address deployment constraints in resource-limited environments.
We first leverage a high-performance Multimodal Large Language Model (MLLM) to generate the initial reasoning dataset and train a medium-sized assistant model with a multi-task learning mechanism. A lightweight student model is jointly trained to perform efficient multimodal sentiment reasoning generation and classification.
Extensive experiments on four datasets demonstrate that MulCoT-RD\footnote{ The code and demo are available via \url{https://github.com/123sghn/MulCoT-RD}.}, with only 3B parameters, achieves strong performance on JMSRC while exhibiting robust generalization and enhanced interpretability.
\end{abstract}

\section{Introduction}

With the proliferation of social media and multimedia content, Multimodal Sentiment Analysis (MSA) has emerged as a critical research area attracting significant academic and industry attention \citet{yang2024large, amiriparian2024muse}. MSA of text-image pairs can be categorized into coarse-grained and fine-grained approaches based on sentiment targets. Coarse-grained MSA \cite{yang2021multimodal, zhang2023learning} identifies the overall sentiment of text-image pairs, while fine-grained MSA, or Multimodal Aspect-Based Sentiment Classification (MASC) \cite{zhou2023aom, wang2024wisdom, yang2025learning}, analyzes sentiment toward specific aspect terms within textual content.

\begin{figure}[t]
\centering
\includegraphics[width=0.48\textwidth]{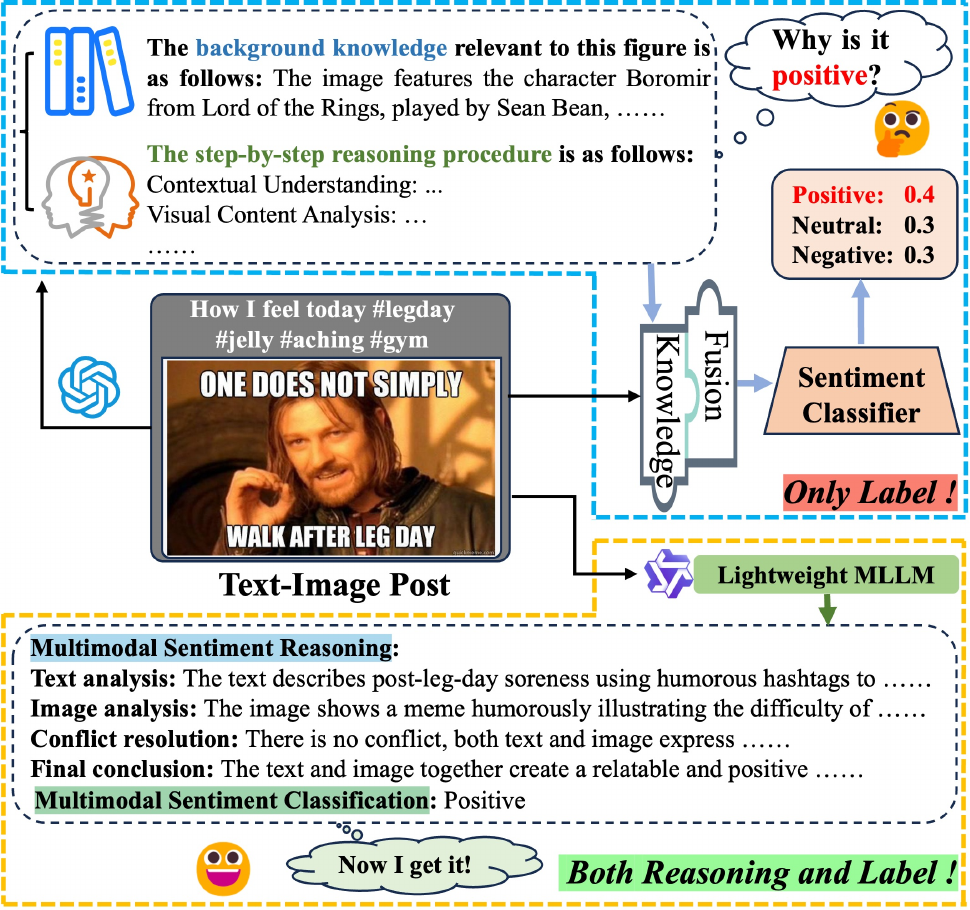}
\caption{Leveraging reasoning (blue dashed line) vs.  Generating reasoning chain (yellow dashed line).}
\label{figure_introduction}
\vspace{-2em}
\end{figure}

Most existing methods enhance MSA through multimodal representation learning \cite{zhang2022affective, manzoor2023multimodality} and fusion \cite{huang2020multimodal, zhang2023learning}, employing separate encoders to extract unimodal representations, then integrating them using fusion strategies such as gating mechanisms \cite{kumar2020gated}, cross-modal attention \cite{ju2021joint}, and graph neural networks \cite{yang2021multimodal}. 
While these approaches advance MSA performance, they face a fundamental limitation: inability to model intra-modal and cross-modal sentiment reasoning processes that explain why users experience particular sentiments. These models typically operate as "black boxes" for sentiment classification, obscuring the specific contributions of each modality and interaction mechanisms in sentiment decisions due to their lack of explicit modeling of sentiment presentation and reasoning chain across modalities.

Building upon LLMs, Multimodal Large Language Models (MLLMs) \cite{hurst2024gpt, wu2024deepseek,bai2025qwen2} demonstrate remarkable performance across diverse multimodal tasks, including recommendation systems \cite{ye2025harnessing}, sentiment analysis \cite{wang2024wisdom}, and mental health assessment \cite{zhang2024psydraw}. As shown in Figure \ref{figure_introduction} (blue box), current methods leverage high-performing MLLMs, like GPT-4o, to inject world knowledge or Chain-of-Thought (CoT) \cite{wei2022chain} reasoning into pre-trained language models for MSA improvement \cite{wang2024wisdom, li2025multimodal}, yet fail to transfer superior reasoning capabilities.
Existing research \cite{li2025small} shows that lightweight MLLMs ($\leq$3B parameters) exhibit limited CoT reasoning capabilities, necessitating reliance on models with superior reasoning abilities. However, closed-source models incur substantial costs, while large-scale MLLMs require extensive computational resources, limiting deployment in resource-constrained environments.
Developing lightweight MLLMs (e.g., 3B parameters) that autonomously generate high-quality multimodal sentiment reasoning while maintaining high MSA performance represents a major challenge, as highlighted in the yellow box of Figure \ref{figure_introduction}.

\begin{figure*}[t]
\centering
\includegraphics[width=0.85\textwidth]{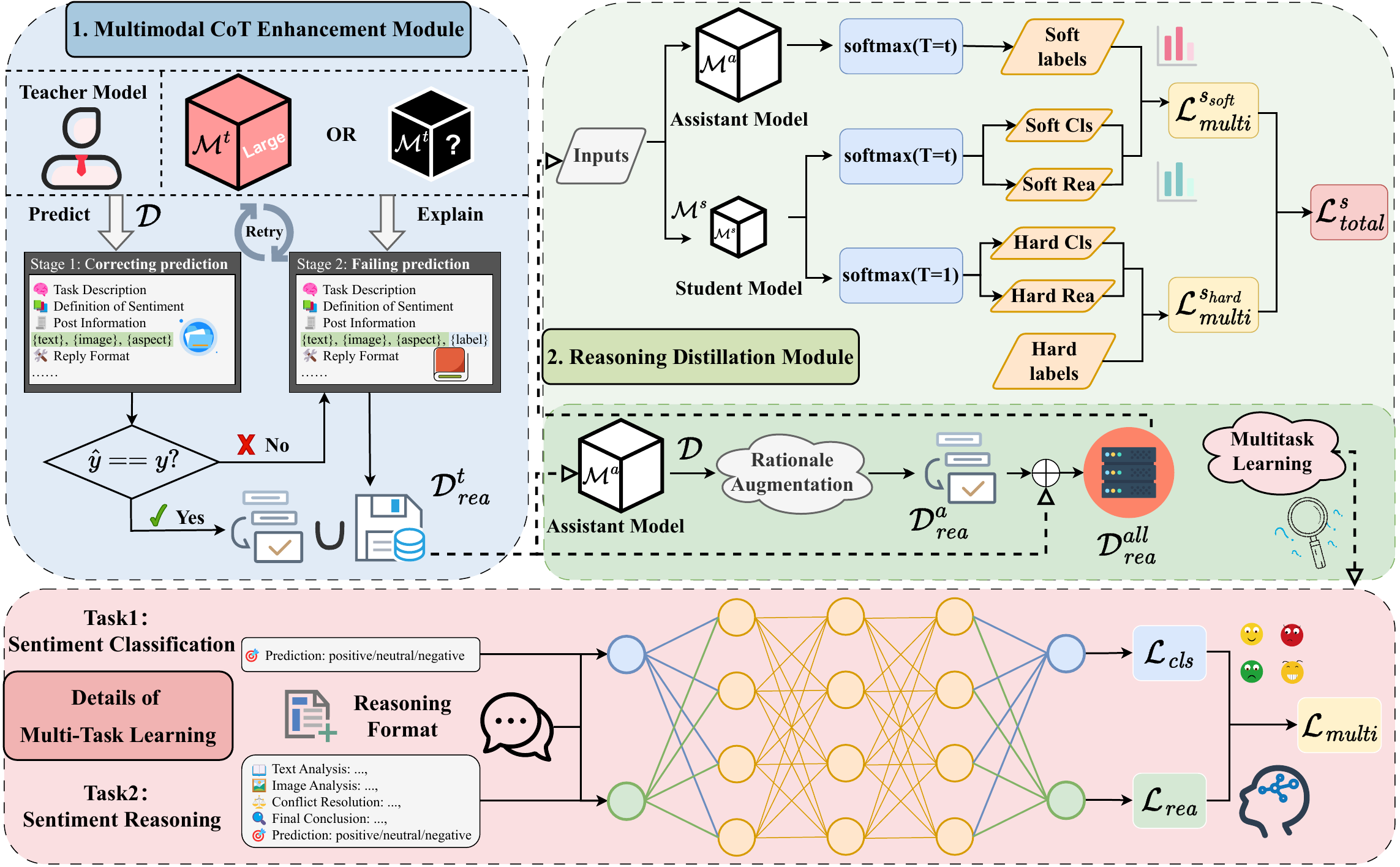}
\caption{MulCoT-RD comprises two core modules, i.e., (1) Multimodal CoT Enhancement Module, (2) Reasoning Distillation Module (Assistant Model with Multi-Task Learning, Student Model with Joint Learning).}
\label{figure_architecture}
\vspace{-1.5em}
\end{figure*}

To address these challenges, we focus on the \textbf{Resource-Limited Joint Multimodal Sentiment Reasoning and Classification} \textbf{(JMSRC)} task, which simultaneously performs multimodal sentiment reasoning generation and classification using only a lightweight MLLM.
We introduce the \textbf{Multimodal Chain-of-Thought Enhancement with Reasoning Distillation (MulCoT-RD)} framework for JMSRC, illustrated in Figure \ref{figure_architecture}, while leveraging Reasoning Distillation (RD) with the Teacher-Assistant-Student pattern to enable lightweight MLLMs to autonomously generate high-quality sentiment reasoning (for the second challenge).
The MulCoT-RD comprises two core modules. \textbf{(1) Multimodal CoT Enhancement Module}: We design a two-stage module using structured prompt templates with task decomposition, reasoning guidance, conflict mediation steps, and adaptive retry control. It guides the high-performance closed-source or large-scale open-source MLLM as a teacher model to generate logically coherent multimodal sentiment reasoning.
\textbf{(2) Multimodal Sentiment Reasoning Distillation Module}: Considering teacher model limitations in providing soft labels and intermediate representations, data scarcity, and inference costs, we introduce a medium-sized open-source MLLM as an assistant model, and use it to synthesize high-quality data. Through multi-task learning, the assistant model jointly enhances sentiment label prediction accuracy and reasoning quality.
For efficient deployment in resource-constrained environments, we employ joint optimization combining hard labels with soft labels from the assistant model to transfer reasoning capabilities to a lightweight student MLLM, achieving optimal balance among classification performance, interpretability, and deployment efficiency.
Our contributions are summarized as follows:

\begin{itemize}
\setlength{\itemsep}{0.2em}  %
\setlength{\parskip}{0pt}    %
\setlength{\topsep}{0.3em}   %
    \item We focus on joint multimodal sentiment reasoning and classification in resource-constrained scenarios and construct a high-quality sentiment reasoning dataset.
    \item We propose the Multimodal Chain-of-Thought Enhancement with Reasoning Distillation, MulCoT-RD, framework for JMSRC. Multi-task learning and joint optimization improve the sentiment classification and reasoning capabilities of the model.
    \item Comprehensive experiments across multiple MSA datasets demonstrate that our lightweight 3B-parameter MLLM achieves superior sentiment classification performance while maintaining high interpretability.
\end{itemize}

\section{Related Work}

\subsection{Multimodal Sentiment Analysis}

The MSA development can be broadly divided into two stages: the era of pre-trained language models (PLMs) and the era of large language models (LLMs).
During the PLMs era, MSA methods typically utilize a dedicated encoder for each modality to extract representations, with a primary focus on multimodal fusion and cross-modal alignment. \cite{zhang2023learning,xiao2023cross,zhou2023aom}. 
The emergence of LLMs has opened new possibilities for MSA. However, existing methods typically rely on MLLMs to generate valuable knowledge \cite{wang2024wisdom} or reasoning \cite{pang2024enhancing, li2025multimodal}, which is then injected into pre-trained language models to improve MSA, rather than enabling autonomous sentiment reasoning. This results in limited interpretability.
To our knowledge, Emotion-LLaMA is the first LLM-based model for multimodal emotion recognition and explanation, but requires modality-specific representation learning, pre-training, and instruction tuning \cite{cheng2024emotion}.
Models with superior reasoning capabilities are often computationally expensive or have large parameter counts that complicate deployment. We focus on using lightweight MLLM to simultaneously achieve efficient and autonomous generation of high-quality multimodal sentiment reasoning and classification.

\subsection{Reasoning Distillation}

Knowledge Distillation (KD) \cite{hinton2015distilling} has proven effective for compressing language models by transferring predictive behaviors, such as soft labels or hidden representations, from larger teacher models to smaller student models. Current KD techniques for PLMs focus on distilling soft labels \cite{sanh2019distilbert, gu2023minillm} or representations \cite{wang2020minilm, wang2020minilmv2, kim2022tutoring}, but require access to the teacher model's internal parameters. This dependency creates significant challenges when applying KD to closed-source LLMs.
Reasoning distillation offers an alternative approach, enabling smaller student models to acquire reasoning capabilities by fine-tuning on reasoning processes from a teacher model instead of relying on soft labels \cite{magister2022teaching, li2023symbolic, lee2024mentor, chenglin2024mixed}. 
In our work, we leverage an intermediate-sized model with multi-task learning as an assistant to both supplement soft-label distillation signals from the teacher model and generate higher-quality data to address reasoning data scarcity.

\section{Method}

To achieve an effective integration of task performance, interpretability, and deployment efficiency, we introduce the Multimodal Chain-of-Thought Enhancement with Reasoning Distillation (MulCoT-RD) framework for JMSRC, as shown in Figure \ref{figure_architecture}, comprising the Multimodal CoT Enhancement Module and the Reasoning Distillation Module.

\subsection{Task Definition}
Given a dataset $\mathcal{D} = \left\{ (x_i, L_i) \right\}_{i=1}^N$ containing $\mathit{N}$ samples, each sample $x_i$ consists of text $T_i$, image $I_i$, aspect term $[A_i]$ (provided only in fine-grained MSA), and sentiment label $L_i$.
The JMSRC task is formulated as follows: 
\begin{equation}
\mathcal{M}\left(T_i,\ I_i,\left[A_i\right]\right)\Rightarrow\left (R_i, \hat{y}_i\right),
\end{equation}
where $R_i$ denotes the corresponding sentiment reasoning, and $\hat{y}_i$ denotes the predicted sentiment label by MLLM $\mathcal{M}$.

\subsection{Multimodal CoT Enhancement}

We propose a two-stage multimodal CoT enhancement module to synthesize high-quality sentiment reasoning data. The corresponding prompts are illustrated in Figure \ref{figure_two_stage_templates} provided in Appendix~\ref{appendix_two_stage_template}.
\textbf{In the first stage}, we perform reasoning path generation in a label-free setting using a high-performance MLLM as the teacher model $\mathcal{M}^t$. We employ a structured CoT prompt template $\mathcal{T}_{pre}$ for \textbf{prediction}, comprising the basic template $\mathcal{T}_{b}$ (including Task Description, Sentiment Definition, and Reasoning Format) and the specific prediction prompt $\mathcal{P}_{pre}$. This template guides the model through text analysis, image analysis, conflict resolution, and conclusion generation, ensuring logically coherent and interpretable reasoning.
\begin{equation}
 c_i^{t_{1}},\hat{y}^t_i =\mathcal{M}^t\left(x_i;\mathcal{T}_{pre}\right),
\end{equation}
where $c_i^{t_1}$ represents the CoT reasoning process generated in the first stage, and $\hat{y}^t_i$ indicates the predicted sentiment label for the $i$-th sample.  

For correctly predicted samples, the generated reasoning paths are directly retained for subsequent training, thereby constructing the first-stage training set, $\mathcal{D}_{rea}^{t1}$.
\begin{equation}
\mathcal{D}_{rea}^{t_1} = \left\{ \left( x_i, c_i^{t_{1}}, \hat{y}^t_i \right) \mid \hat{y}^t_i = L_i \right\}_{i=1}^{N_{t_1}}.
\end{equation}

Misclassified samples often reflect complex cases with ambiguous boundaries or cross-modal conflicts, or semantic ambiguity. Guiding the model to learn causally consistent reasoning on these challenging examples can enhance its understanding and robustness in complex scenarios. Therefore, we design a second stage where, for samples with incorrect predictions, the ground truth label, $L_i$, is introduced and an explain template, $\mathcal{T}_{exp}$, is constructed to guide the model in generating a supervised reasoning process, $c_i^{t_2}$, conditioned on the correct label. 
\begin{equation}
\left\{
\begin{array}{l}
 c_i^{t_{2}}, \hat{y}^t_i = \mathcal{M}^t\left( x_i, L_i; \mathcal{T}_{exp} \right) \\
\mathcal{D}^{t_2}_{rea} = \left\{ \left( x_i, c_i^{t_{2}},  L_i \right) \right\}_{i=1}^{N_{t_2}},
\end{array}
\right.
\end{equation}
where $N=N_{t1} + N_{t2}$; $\mathcal{T}_{exp}$ is constructed by the basic template, $\mathcal{T}_{b}$, and the specific reasoning prompt, $\mathcal{P}_{exp}$, as shown in Figure \ref{figure_two_stage_templates} provided in Appendix~\ref{appendix_two_stage_template}.

The two-stage datasets are merged to obtain the reasoning dataset $\mathcal{D}^t_{rea} = \mathcal{D}^{t_1}_{rea} \cup \mathcal{D}^{t_2}_{rea}$. To improve sentiment reasoning and label prediction reliability, we introduce an adaptive retry controller, ARC, that automatically regenerates outputs when MLLMs produce incomplete structures or invalid labels until a valid result is obtained or the retry limit is reached, ensuring generation quality while controlling computational overhead. The details of ARC are provided in Appendix~\ref{appendix_arc}.

\subsection{Multimodal Sentiment Reasoning Distillation}

Closed-source teacher models limit knowledge extraction due to restricted intermediate representations, while open-source models with strong reasoning often require large parameters \cite{li2025small}, hindering efficient deployment. To address multimodal sentiment reasoning data scarcity and the absence of soft labels, we introduce reasoning distillation \cite{lee2024mentor} to train an \textbf{assistant model with multi-task learning}, as illustrated in the middle right of Figure \ref{figure_architecture}, enhancing data diversity. A \textbf{student model with joint learning}, as shown in the upper right of Figure \ref{figure_architecture}, adapts to resource-constrained environments while inheriting the assistant model's sentiment reasoning and classification capabilities.

\subsubsection{Assistant Model with Multi-Task Learning} We propose a multi-task learning framework that shares hard parameters to train the assistant model, $\mathcal{M}^a$, for JMSRC that jointly optimizes two complementary tasks, including multimodal sentiment reasoning and classification, as shown in the lower part of Figure \ref{figure_architecture}.
 \begin{align}
\mathcal{L} = \frac{-1}{B} \sum_{i=1}^{B} \sum_{j=1}^{l}
& \log P \Big( y_j^{(i)} \mid y_{<j}^{(i)}, \mathcal{M}^a(x^{(i)}) \Big) \notag \ \\
& \cdot I_{{y_j^{(i)} \neq -100}},
\end{align}
where $B$ denotes the batch size; $l$ denotes the target sequence length of the $i$-th sample; $P$ denotes the predicted probability of $y_j^{(i)}$ at decoding step $j$ based on $y_{<j}^{(i)}$; $I_{{ y_j^{(i)} \neq -100 }}$ indicates that only tokens whose labels are not equal to -100 (i.e., not masked) participate in the loss.

The overall loss function for training the assistant model is formulated as follows:
\begin{equation}
    \mathcal{L}^{a}_{multi} = \lambda^{a}_{cls} \cdot \mathcal{L}^{a}_{cls} + \lambda^{a}_{rea} \cdot \mathcal{L}^{a}_{rea},
\label{eq:asst_mtl_loss}
\end{equation}
where $\lambda^{a}_{cls}$ and $\lambda^{a}_{rea}$ are the weighting hyperparameters to ensure a balanced trade-off between two tasks. After training, we can obtain the trained assistant model, $\overline{M}^a$.

Regarding data augmentation, given the limited capabilities of the assistant model, we only retain training samples for which sentiment can be correctly predicted through sentiment reasoning. See the material for more details.
\begin{equation}
    \mathcal{D}^{a}_{rea}=\{\left(x_i,\widehat{{c}}^a_i,\widehat{y}^a_i\right)\mid
    \widehat{y}^a_i=L_i\}_{i=1}^{N_a},
\end{equation}
where $ \widehat{c}^a_i,\widehat{y}^a_i=   \overline{\mathcal{M}}^{a}\left(x_i;\mathcal{T}_{\mathrm{pre}}\right)$ and $N_a < N$.

The complete sentiment reasoning dataset is obtained, which is used to train a student model.
\begin{equation}
    \mathcal{D}^{all}_{rea}=\mathcal{D}^t_{rea}\cup \mathcal{D}^{a}_{rea}.
\label{equation8}
\end{equation}

\subsubsection{Student Model with Joint Learning}

To enable efficient deployment in resource-constrained environments, we employ a lightweight student MLLM, $\mathcal{M}^s$, trained through knowledge distillation. 
The student model jointly learns from two sources, including ground-truth labels (hard labels) for accurate prediction and probability distributions (soft labels) from the assistant model to capture its reasoning patterns. The dual supervision allows the student model to inherit the assistant model's discriminative capabilities.

\textbf{Hard Label.} The student model undergoes fine-tuning using constructed reasoning data, $\mathcal{D}^{all}_{rea}$, enabling it to acquire step-by-step reasoning capabilities through reasoning distillation. The hard label loss is defined as follows:

\begin{equation}
\begin{cases}
\mathcal{L}^{s_{hard}}_{cls} = \mathbb{E}_{\mathcal{D}_{rea}^{all}} \log P \left( [x; L] \mid \mathcal{M}^s \right)  \\
\mathcal{L}^{\mathrm{s}_{hard}}_{rea} = \mathbb{E}_{\mathcal{D}_{rea}^{all}} \log P\left( [x; c] \mid \mathcal{M}^s \right), 
\end{cases}
\end{equation}
   
where $P$ denotes the probability distribution; $c$ represents the reasoning process. The losses $\mathcal{L}^{s_{hard}}_{cls}$ and $\mathcal{L}^{s_{hard}}_{rea}$ are used to train the student model to learn the direct mapping from multimodal input to sentiment labels and to generate coherent sentiment reasoning, respectively.

\textbf{Soft Label.} To address the black-box nature of closed-source MLLMs, the assistant model is employed as an intermediary to provide soft labels for distillation. Given an input \( x \), the probability distribution \( p_k \) at the \( k \)-th position is obtained from the logit value \( z_k \) through a single forward pass followed by the softmax function. It is formally defined as:
\begin{equation}
    p_k=\frac{\exp{\ \left(\ z_k/\tau\ \right)}}{\sum_{j}exp\left(\ z_j/\tau\ \right)},
\end{equation}
where \( \tau \) denotes the temperature hyperparameter, which is used to control the smoothness of the distribution.

After obtaining the probability distributions \( p^a \) from $\mathcal{M}^a$ and \( p^s \) from $\mathcal{M}^s$, we employ the Kullback–Leibler (KL) \cite{wu2025rethinking} divergence to minimize the discrepancy between the two distributions. It enables the student model to mimic the prediction behavior of the larger model. The training for soft label distillation is defined as follows:
\begin{equation}
\begin{cases}
\mathcal{L}_{soft}\left(p^a,p^s\right)=\sum_{k} p_k^a\log{\frac{p_k^a}{p_k^s}}\\
\mathcal{L}^{s_{soft}}_{cls}=\mathcal{L}_{soft}\left(p^a_{cls},p^s_{cls}\right) \\
\mathcal{L}^{s_{soft}}_{rea}=\mathcal{L}_{soft}\left(p^a_{rea},p^s_{rea}\right). 
\end{cases}
\end{equation}

\textbf{Joint Learning.} The student model training retains the multi-task learning. The overall hard-label loss and soft-label loss for the student model are defined as follows:
\begin{equation}
\begin{cases}
    \mathcal{L}^{s_{hard}}_{multi} = \lambda^{s_{hard}}_{cls} \cdot \mathcal{L}^{s_{hard}}_{cls} + \lambda^{s_{hard}}_{rea} \cdot \mathcal{L}^{s_{hard}}_{rea}\\
    \mathcal{L}^{s_{soft}}_{multi} = \lambda^{s_{soft}}_{cls} \cdot \mathcal{L}^{s_{soft}}_{cls} + \lambda^{s_{soft}}_{rea} \cdot \mathcal{L}^{s_{soft}}_{rea}
\end{cases}
\end{equation}
where $\lambda^{s_{hard}}_{cls}$, $\lambda^{s_{hard}}_{rea}$, $\lambda^{s_{soft}}_{cls}$, and $\lambda^{s_{soft}}_{rea}$ are hyperparameters that balance the contributions of classification loss and reasoning generation loss in the hard-label and soft-label multi-task learning objectives, respectively.

To jointly leverage hard-label and soft-label supervision, we define the total loss of the student model as follows.
\begin{equation}
    \mathcal{L}^{s}_{total}=\left(1-\lambda\right)\mathcal{L}^{s_{hard}}_{multi}+\lambda\mathcal{L}^{s_{soft}}_{multi},
\end{equation}
where $\lambda$ is a hyperparameter that controls the balance between hard-label and soft-label supervision.

\section{Experiments}

\subsection{Experimental Settings}

\subsubsection{Datasets}

We conduct experiments on both coarse-grained MSA, MVSA-Single and MVSA-Multiple datasets, preprocessed following \cite{liu2024weakly} and fine-grained MSA, Twitter-2015 and Twitter-2017 datasets \cite{yu2019adapting}. Table~\ref{table_statistics} presents the statistics of four datasets with the constructed sentiment reasoning data for JMSRC.
\begin{table}[h]
\centering
\small  %
\setlength{\tabcolsep}{3pt}  %
\begin{tabular}{c c c c 
    >{\columncolor{gray!20}}c %
    >{\columncolor{gray!40}}c  %
}
\toprule[1pt]
\textbf{Dataset} & \textbf{Train} & \textbf{Dev} & \textbf{Test} & \textbf{Train$^{g+}$} & \textbf{Train$^{q+}$} \\
\midrule
\textbf{MVSA-Single} & 3608 & 451 & 452 &  6483 & 6350 \\
\textbf{MVSA-Multiple} & 13619 & 1702 & 1702 & 23424 & 23697 \\
\textbf{Twitter-2015} & 3179 & 1122 & 1037 & 6166 & 6218 \\
\textbf{Twitter-2017} & 3562 & 1176 & 1234 & 6652 & 6871 \\
\bottomrule[1pt]
\end{tabular}
\caption{Statistics of datasets. ${g+}$ and ${q+}$ represent the teacher models GPT-4o-mini \cite{hurst2024gpt} and Qwen2.5-VL-72B \cite{bai2025qwen2}, respectively.}
\label{table_statistics}
\end{table}

\subsubsection{Model Selection}

To build an efficient hierarchical reasoning distillation, we design four distillation architectures, as summarized in Table~\ref{table_four_architectures}. Note that, while our model selection is limited, experimental results clearly demonstrate the effectiveness of MulCoT-RD. See the Appendix~\ref{appendix_model_selection} for more details.

\begin{table}[h]
\centering
\small  %
\setlength{\tabcolsep}{2.5pt}  %
\begin{tabular}{c c c c}
\toprule
\textbf{ID} & \textbf{Teacher Model} & \textbf{Assistant Model} & \textbf{Student Model} \\
\midrule
1 & \multirow{2}{*}{\textbf{GPT-4o-mini}} & Qwen3-VL-8B     & Qwen3-VL-2B \\
2  &                                      & Qwen2.5-VL-7B   & Qwen2.5-VL-3B \\
3 & \multirow{2}{*}{\textbf{Qwen2.5-VL-72B}} & Qwen3-VL-8B  & Qwen3-VL-2B \\
4  &                                      & Qwen2.5-VL-7B   & Qwen2.5-VL-3B \\
\bottomrule
\end{tabular}
\caption{Four reasoning distillation architectures.}
\label{table_four_architectures}
\vspace{-2.0em}
\end{table}

\subsubsection{Implementation Details}

We train our models on NVIDIA RTX A6000 GPUs using the AdamW optimizer \cite{loshchilov2017decoupled}. During training, we set the initial learning rate to 3e-4 and employ a dynamic adjustment strategy: if the validation set performance does not improve for two consecutive epochs, we halve the learning rate until it reaches a minimum of 1e-6. Due to resource limitations, we set the batch size to 2 and train for a maximum of 20 epochs. To mitigate instability caused by small batch sizes, we use gradient accumulation, updating parameters every 20 steps. The multi-task learning hyperparameters {$\lambda^{a}_{rea}$, $\lambda^{s_{hard}}_{rea}$, $\lambda^{s_{soft}}_{rea}$} and {$\lambda^{a}_{cls}$, $\lambda^{s_{hard}}_{cls}$, $\lambda^{s_{soft}}_{cls}$} are set to 0.8 and 0.2, respectively, while the knowledge distillation coefficient $\lambda$ is set to 0.3. Detailed explanations and configurations can be found in Appendix~\ref{appendix_implementation_details}

\subsubsection{Evaluation Metrics}

In line with previous work \cite{chen2024d2r}, we evaluate model performance of classification on coarse-grained MSA using Accuracy (\textbf{Acc}) and Weighted F1 (\textbf{w-F1}). For fine-grained MSA (MASC), we follow previous studies \cite{zhou2023aom} and adopt Accuracy and Macro F1 (\textbf{m-F1}) as evaluation metrics.
For the sentiment reasoning task, we employ comprehensive metrics including sentence embedding-based cosine similarity (\textbf{Sim}) \cite{reimers2019sentence}, \textbf{METEOR} \cite{banerjee2005meteor}, \textbf{BLEU} \cite{papineni2002bleu}, \textbf{ROUGE-L} \cite{lin2004rouge}, and Distinct-N1/N2 (\textbf{Dist-1/2}) \cite{li2015diversity}. 

\subsection{Baselines}

We compare popular models on \textbf{coarse-grained MSA} with MulCoT-RD, including \textbf{MultiSentiNet} \cite{xu2017multisentinet}, \textbf{HSAN} \cite{xu2017analyzing}, \textbf{CoMN-Hop6} \cite{xu2018co}, \textbf{MGNNS} \cite{yang2021multimodal},  \textbf{CLMLF} \cite{li2022clmlf}, \textbf{MVCN} \cite{wei2023tackling}, \textbf{$\mathbf{D}^2 \mathbf{R}$} \cite{chen2024d2r}.
For \textbf{fine-grained MSA}, involving \textbf{ESAFN} \cite{yu2019entity}, \textbf{TomBERT} \cite{yu2019adapting}, \textbf{CapTrBERT} \cite{khan2021exploiting}, \textbf{JML} \cite{ju2021joint}, \textbf{VLP-MABSA} \cite{ling2022vision}, \textbf{CMMT} \cite{yang2022cross}, \textbf{AoM} \cite{zhou2023aom}, \textbf{AETS} \cite{zhu2025aspect}.
\textbf{Emotion-LLaMA} \cite{cheng2024emotion} employs pretraining and instruction tuning based on LLaMA2-7B-Chat to enhance multimodal emotion recognition and explanation. \textbf{Qwen3-VL-8B-Thinking} \cite{yang2025qwen3} features Interleaved-MRoPE and DeepStack for powerful spatial-temporal reasoning. Detailed descriptions can be found in Appendix~\ref{appendix_baselines}.

\subsection{Main Results}
Unlike previous models that only perform multimodal sentiment classification, our model enables joint sentiment reasoning and classification. We conduct experiments on both multimodal sentiment classification and reasoning tasks.

\subsubsection{Results of Multimodal Sentiment Classification}

\textbf{Performance on coarse-grained MSA.} Table~\ref{table_results_MSA} presents the comparison results on the coarse-grained MSA task. MulCoT-RD outperforms both the second-best model (Emotion-LLaMA) and the previous state-of-the-art model ($\mathbf{D}^2 \mathbf{R}$) on the MVSA-Single and MVSA-Multiple datasets, achieving substantial improvements. It highlights the benefits of explicitly modeling intra-modal sentiment structures and cross-modal reasoning processes. Notably, although the teacher model has greater parameter capacity, its lack of task-specific fine-tuning for MSA leads to suboptimal modeling of cross-modal emotional relations, making it inferior to the assistant model optimized with task-oriented objectives. Moreover, the student model outperforms the assistant model in certain cases, likely due to benefiting from the augmented training data generated by the assistant, which improves its generalization and robustness.

\begin{table}[h]
\centering
\small
\setlength{\tabcolsep}{3pt}
\begin{tabular}{c| c| c c c c}
\toprule
\multirow{2}{*}{\textbf{Model}} & \multirow{2}{*}{\textbf{Venue}} & \multicolumn{2}{c}{\textbf{MVSA-S}} & \multicolumn{2}{c}{\textbf{MVSA-M}} \\
 & &  \textbf{Acc} & \textbf{w-F1} & \textbf{Acc} & \textbf{w-F1} \\
\midrule
MultiSentiNet & CIKM'17 & 69.8 & 69.8 & 68.9 & 68.1 \\
HSAN & ISI'17 & 69.9 & 66.9 & 68.0 & 67.8 \\
CoMN-Hop6 & SIGIR'18 & 70.5 & 70.0 & 68.9 & 68.8 \\
MGNNS & ACL'21 & 73.8 & 72.7 & 72.5 & 69.3 \\
CLMLF & NAACL'21 & 75.3 & 73.5 & 72.0 & 69.8 \\
MVCN & ACL'23 & 76.1 & 74.6 & 72.1 & 70.0 \\
{$\mathbf{D}^2 \mathbf{R}$} & EMNLP'24 & 76.7 & 75.6 & 71.6 & 70.9 \\
\midrule
Emotion-LLaMA$^{\dagger}$ & NeurIPS'24 & 82.7 & 81.8 & 75.6 & \textbf{75.2} \\
Open-Flamingo$^{\ddagger}$ & ICML'25 & 66.3 & - & 68.7 & - \\
Qwen2.5-VL-3B$^*$ & Student & 62.8 & 66.4 & 74.2 & 70.7 \\
Qwen2.5-VL-7B$^*$ & Assistant & 67.7 & 69.6 & 74.7 & 70.9 \\
3-VL-8B-Thinking$^*$ &  & 72.3 & 71.5 & 70.2 & 69.4 \\
\rowcolor{gray!20}
GPT-4o-mini$^*$ & Teacher$^1$ & 76.7 & 75.6 & 71.6 & 71.4 \\
\rowcolor{gray!20}
\textbf{MulCoT-RD(asst)} &  & \textbf{83.6} & \underline{82.8} & 75.7 & 72.9 \\
\rowcolor{gray!20}
\textbf{MulCoT-RD(stu)} &  & 82.7 & 82.3 & \underline{76.9} & 74.2 \\
\rowcolor{gray!40}
Qwen2.5-VL-72B$^*$ & Teacher$^2$ & 67.9 & 70.8 & 74.2 & 71.8 \\
\rowcolor{gray!40}
\textbf{MulCoT-RD(asst)} &  & 83.2 & 82.1 & \underline{76.9} & 73.8 \\
\rowcolor{gray!40}
\textbf{MulCoT-RD(stu)} &  & \underline{83.4} & \textbf{83.2} & \textbf{77.2} & \underline{74.4} \\
\bottomrule
\end{tabular}
\caption{Results for coarse-grained MSA. Models above the middle line are small models fully fine-tuned, while those below are (M)LLMs fine-tuned with LoRA. $^{\dagger}$ denotes the results reproduced by us using models retrained on our datasets. $^{\ddagger}$ indicates the 16-shot performance under In-Context Learning (ICL). The best results are bold-typed and the second best ones are underlined. $^{*}$ means the zero-shot performance.}
\label{table_results_MSA}
\vspace{-2.0em}
\end{table}

\textbf{Performance on MASC.} As shown in Table~\ref{table_results_MASC}, the MulCoT-RD(asst) model (with Qwen2.5-VL-72B as the teacher) achieves the best overall performance. Compared to the second-best models AoM and AETS, MulCoT-RD(asst) exhibits a slight decrease in accuracy on the Twitter-2017 dataset by 1.4\% and 1.6\%, respectively, but consistently achieves the highest scores across all other evaluation metrics. We attribute this to two primary reasons. First, the Twitter-2017 dataset contains a large number of unparseable and unrecognizable symbols \cite{peng2024novel}, including emojis that are commonly used on Twitter. These symbols may mislead the model by obscuring emotional semantics during reasoning, thereby slightly reducing accuracy. Second, MulCoT-RD(asst) is fine-tuned using LoRA, whereas most existing SOTA methods, such as AoM and AETS, adopt full-parameter fine-tuning. This limits the extent of parameter updates during task adaptation, resulting in smaller performance gains compared to full fine-tuning \cite{biderman2024lora}. Given this, we believe our proposed method remains effective for MASC.

Notably, the student model of MulCoT-RD contains only 3B parameters, significantly fewer than the large multimodal architecture of Emotion-LLaMA \cite{cheng2024emotion}, which combines LLaMA2-7B-chat with encoders like EVA, CLIP, VideoMAE, and HuBERT-large. Despite its smaller size, MulCoT-RD(stu) outperforms Emotion-LLaMA on multiple benchmarks, demonstrating superior efficiency and strong applicability in resource-constrained settings.

\begin{table}[h]
\centering
\small
\setlength{\tabcolsep}{2.5pt}
\begin{tabular}{c| c| c c c c}
\toprule[1pt]
\multirow{2}{*}{\textbf{Model}} & \multirow{2}{*}{\textbf{Venue}} & \multicolumn{2}{c}{\textbf{Twitter-15}} & \multicolumn{2}{c}{\textbf{Twitter-17}} \\
&      & \textbf{Acc} & \textbf{m-F1} & \textbf{Acc} & \textbf{m-F1} \\
\midrule
ESAFN & TASLP'20 & 73.4 & 67.4 & 67.8 & 64.2 \\
TomBERT & IJCAI'19 & 77.2 & 71.8 & 70.5 & 68.0 \\
CapTrBERT & ACM MM'21 & 78.0 & 73.2 & 72.3 & 70.2 \\
JML & EMNLP'21 & 78.7 & - & 72.7 & - \\
VLP-MABSA & ACL'22 & 78.6 & 73.8 & 73.8 & 71.8 \\
CMMT & IPM'22 & 77.9 & - & 73.8 & - \\
AoM & ACL'23 & 80.2 & \underline{75.9} & \underline{76.4} & \underline{75.0} \\
AETS & AAAI'25 & 79.5 & - & \textbf{76.6} & - \\
\midrule
Emotion-LLaMA$^{\dagger}$ & NeurIPS'24 & 73.9 & 70.2 & 69.2 & 67.9 \\
Open-Flamingo$^{\ddagger}$ & ICML'25 & 70.4 & - & 62.6 & - \\
Qwen2.5-VL-3B$^*$ & Student & 48.9 & 49.7 & 56.8 & 55.6 \\
Qwen2.5-VL-7B$^*$ & Assistant & 58.3 & 55.6 & 58.6 & 57.6 \\
3-VL-8B-Thinking$^*$ &  & 59.2 & 55.8 & 60.4 & 58.9 \\
\rowcolor{gray!20}
GPT-4o-mini$^*$ & Teacher$^1$ & 49.4 & 37.6 & 54.0 & 52.8 \\
\rowcolor{gray!20}
\textbf{MulCoT-RD(asst)} &  & \underline{80.7} & 75.3 & 74.6 & 74.6 \\
\rowcolor{gray!20}
\textbf{MulCoT-RD(stu)} &  & 80.4 & 75.2 & 74.0 & 73.3 \\
\rowcolor{gray!40}
Qwen2.5-VL-72B$^*$ & Teacher$^2$ & 59.5 & 57.1 & 63.9 & 63.4 \\
\rowcolor{gray!40}
\textbf{MulCoT-RD(asst)} &  & \textbf{80.8} & \textbf{77.2} & 75.0 & \textbf{75.1} \\
\rowcolor{gray!40}
\textbf{MulCoT-RD(stu)} &  & 80.5 & 75.1 & 74.3 & 74.1 \\
\bottomrule[1pt]
\end{tabular}
\caption{Results of different methods for MASC. ``-'' means it does not exist in the original paper.}
\label{table_results_MASC}
\vspace{-2.0em}
\end{table}

\subsubsection{Evaluation of Sentiment Reasoning} 
MulCoT-RD achieves efficient and effective sentiment reasoning. We evaluate the reasoning performance of the student and assistant models, as well as Emotion-LLaMA, using the sentiment reasoning process from the teacher model as gold-standard references (exemplified by GPT-4o-mini), with results presented in Table~\ref{table_cot_evaluation}.
Our models achieve a comprehensive performance advantage over Emotion-LLaMA across all key reasoning metrics. The results demonstrate high-quality sentiment reasoning generation across multiple evaluation metrics. Cosine similarity (Sim) consistently exceeds 90\% across all models, confirming strong semantic alignment between generated and gold-standard reasoning chains. METEOR scores ranging from 45.4\% to 59.8\% further indicate substantial paraphrase-level and lexical overlap. While BLEU and ROUGE-L show some fluctuations, coarse-grained MSA variants generally outperform fine-grained MSA, reflecting better surface-form alignment. Distinct-N1 and Distinct-N2 scores remain approximately 49\% and 80\%, respectively, indicating that the generated reasoning maintains high linguistic diversity, enhancing the interpretability and robustness of reasoning tasks. 
\begin{table}[h]
\centering
\small
\setlength{\tabcolsep}{1.5pt}
\begin{tabular}{c| c| c c c c c c}
\toprule[1pt]
{\textbf{Model}} & \textbf{Dataset} & \textbf{Sim} & \textbf{Meteor} & \textbf{Bleu} & \textbf{Rouge-L} & \textbf{Dist-1} & \textbf{Dist-2} \\
\midrule
\multirow{4}{*}{\textbf{ELLA}} & MVSA-S & 87.6 & 35.9 & 14.6 & 35.1 & 49.8 & 80.2 \\
 & MVSA-M & 84.7 & 36.0 & 15.9 & 35.9 & 52.5 & 83.7 \\
 & Twitter-15 & 86.3 & 38.6 & 18.3 & 39.3 & 42.7 & 72.9 \\
 & Twitter-17 & 86.6 & 38.1 & 17.6 & 38.2 & 43.0 & 73.1 \\
\midrule
\multirow{4}{*}{\textbf{Asst}} & MVSA-S & 92.6 & 59.8 & 47.8 & 55.0 & 49.8 & 80.2 \\
 & MVSA-M & 93.0 & 57.4 & 48.1 & 57.2 & 48.6 & 79.4 \\
 & Twitter-15 & 92.9 & 54.6 & 43.0 & 58.3 & 42.4 & 72.9 \\
 & Twitter-17 & 90.5 & 51.2 & 35.9 & 53.3 & 45.2 & 74.1 \\
 \midrule
\multirow{4}{*}{\textbf{Stu}} & MVSA-S & 92.2 & 47.3 & 58.8 & 54.2 & 49.8 & 80.2 \\
 & MVSA-M & 92.1 & 56.8 & 46.7 & 55.8 & 49.5 & 80.3 \\
 & Twitter-15 & 90.3 & 45.4 & 28.2 & 46.0 & 49.5 & 79.9 \\
 & Twitter-17 & 90.0 & 49.2 & 33.1 & 50.8 & 45.2 & 74.1 \\
\bottomrule
\end{tabular}
\caption{Evaluation results of generated reasoning from Emotion-LLaMA, assistant and student models.}
\label{table_cot_evaluation}
\vspace{-1.5em}
\end{table}

\subsection{Ablation Study}

In this section, we investigate the impact of each MulCoT-RD component, with results presented in Table~\ref{table_results_ablation}.
When we only use the text modality (\textbf{w/o Img}), the model performs worse on all metrics compared to the complete model, highlighting the importance of incorporating visual modality. Similarly, when we remove the text modality (\textbf{w/o Text}), the model has a significant performance drop on all datasets. The decline, more severe than w/o Img, highlights the key role of text and the necessity of multimodal integration.
\textbf{w/o CoT} means to remove the multi-task learning paradigm and exclude the sentiment reasoning task from the training process, leading to a general performance drop. It highlights the importance of deeply modeling intra-modal and cross-modal sentiment reasoning.
Note that all the above ablation experiments are conducted on the assistant model.
\textbf{w/o Asst} omits the assistant model, removing the use of soft labels in the distillation process and reducing the scale and diversity of training data. This leads to a notable performance drop across all datasets, demonstrating the effectiveness of the teacher–assistant–student hierarchical distillation framework for JMSRC.

\begin{table}[h]
\centering
\small
\setlength{\tabcolsep}{2pt}
\begin{tabular}{c|cc|cc|cc|cc}
\toprule
\multirow{2}{*}{\textbf{Method}} & \multicolumn{2}{c|}{\textbf{MVSA-S}} & \multicolumn{2}{c|}{\textbf{MVSA-M}} & \multicolumn{2}{c|}{\textbf{Twitter-15}} & \multicolumn{2}{c}{\textbf{Twitter-17}} \\
& \textbf{Acc} & \textbf{w-F1} & \textbf{Acc} & \textbf{w-F1} & \textbf{Acc} & \textbf{m-F1} & \textbf{Acc} & \textbf{w-F1} \\
\midrule
MulCoT-RD    & \textbf{83.2} & \textbf{82.1} & \textbf{76.9} & 73.8 & \textbf{80.8} & \textbf{77.2} & \textbf{75.0} & \textbf{75.1} \\
\midrule
w/o Img      & 79.4 & 77.7 & 73.7 & 73.0 & 78.4 & 72.5 & 73.5 & 73.5 \\
w/o Txt      & 77.9 & 77.1 & 66.2 & 67.7 & 65.6 & 56.6 & 64.6 & 59.4 \\
w/o CoT      & 79.9 & 79.7 & 74.2 & 73.1 & 79.9 & 75.5 & 74.2 &  73.4\\
w/o Asst     & 81.9 & 81.3 & 75.2 & \textbf{74.1} & 79.3 & 72.3 & 73.7 & 73.3 \\
\bottomrule
\end{tabular}
\caption{The performance comparison of our full model and its ablated methods under the setting where Qwen2.5-VL-72B serves as the teacher model.}
\label{table_results_ablation}
\vspace{-2.5em}
\end{table}

\subsection{Efficiency of MulCoT-RD}

\begin{figure}[h]
\centering
\includegraphics[width=0.48\textwidth]{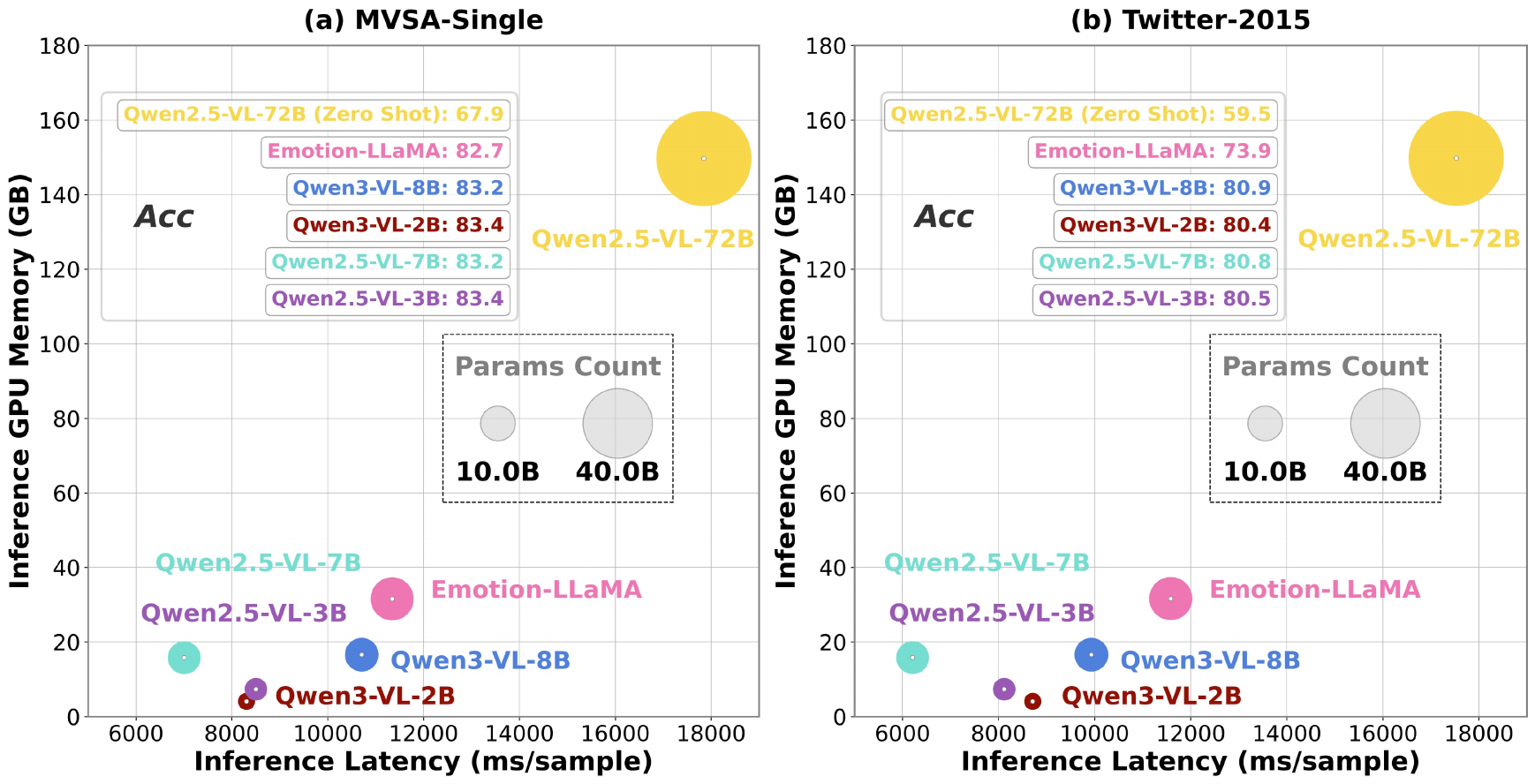}
\caption{Efficiency comparison on MVSA-Single and Twitter-2015. Metrics are measured on the batch size as 1 and all samples are from the test set. Note that the models are trained under the paradigm where Qwen2.5-VL-72B serves as the teacher model. Detailed efficiency comparison results are provided in Appendix~\ref{appendix_efficiency}.}
\label{figure_efficiency}
\vspace{-2.0em}
\end{figure}

To further demonstrate the practicability of MulCoT-RD, we provide the model efficiency comparison in Figure~\ref{figure_efficiency}. We find that, on the MVSA-Single and Twitter-2015 datasets, our distilled student models (Qwen2.5-VL-3B and Qwen3-VL-2B) achieve significantly lower inference latency and GPU memory usage than Emotion-LLaMA, while obtaining notable improvements in Accuracy. Specifically, in Twitter-2015, Qwen3-VL-2B achieves the 8.79\% accuracy improvement (80.4 vs. 73.9) with 5.81x fewer parameters and 0.33x faster inference speed compared to Emotion-LLaMA.

\subsection{Robustness of MulCoT-RD}

To validate the robustness of our approach across different backbones, we conducted the base-model adaptation study by replacing the Qwen2.5-VL series with the Qwen3-VL series (as shown in Table~\ref{table_results_qwen3}) and the Flan-T5 series (see Appendix~\ref{appendix_robustness}). The results demonstrate that the models maintain strong performance across different backbones, illustrating the robustness and adaptability of MulCoT-RD.

\begin{table}[h]
\centering
\small
\setlength{\tabcolsep}{1.5pt}
\begin{tabular}{c|c c c c c c c c}
\toprule
\multirow{2}{*}{\textbf{Qwen3-VL}} & \multicolumn{2}{c}{\textbf{MVSA-S}} & \multicolumn{2}{c}{\textbf{MVSA-M}} & \multicolumn{2}{c}{\textbf{Twitter-15}} & \multicolumn{2}{c}{\textbf{Twitter-17}} \\
 & \textbf{Acc} & \textbf{w-F1} & \textbf{Acc} & \textbf{w-F1} & \textbf{Acc} & \textbf{m-F1} & \textbf{Acc} & \textbf{m-F1}\\
\midrule
\textbf{8B (asst)$^1$} & \underline{83.6} & \underline{82.9} & 76.0 & 73.5 & \underline{80.5} & 76.3 & \textbf{76.1} & \textbf{75.1}\\
\textbf{2B (stu)}$^1$ & \textbf{84.1} & \textbf{83.7} & \underline{76.7} & \underline{74.1} & 80.0 & 75.2 & 74.8 & \underline{74.7}\\
 \midrule
\textbf{8B (asst)}$^2$ & 83.2 & 82.6 & \underline{76.7} & 73.2 & \textbf{80.9} & \textbf{77.4} & \underline{75.0} & 73.4\\
\textbf{2B (stu)}$^2$ & 83.4 & 82.4 & \textbf{76.8} & \textbf{74.2} & 80.4 & \underline{76.7} & 73.8 & 73.0\\
\bottomrule
\end{tabular}
\caption{Performance of Qwen3-VL-based models on coarse-grained MSA and MASC. The best results are bold-typed, and the second best ones are underlined. $1$ and $2$ indicate the Teacher models, with $1$ being GPT-4o-mini and $2$ being Qwen2.5-VL-72B.}
\label{table_results_qwen3}
\vspace{-2.0em}
\end{table}

\subsection{Case Study}

\begin{figure*}[t]
\centering
\includegraphics[width=1.0\textwidth]{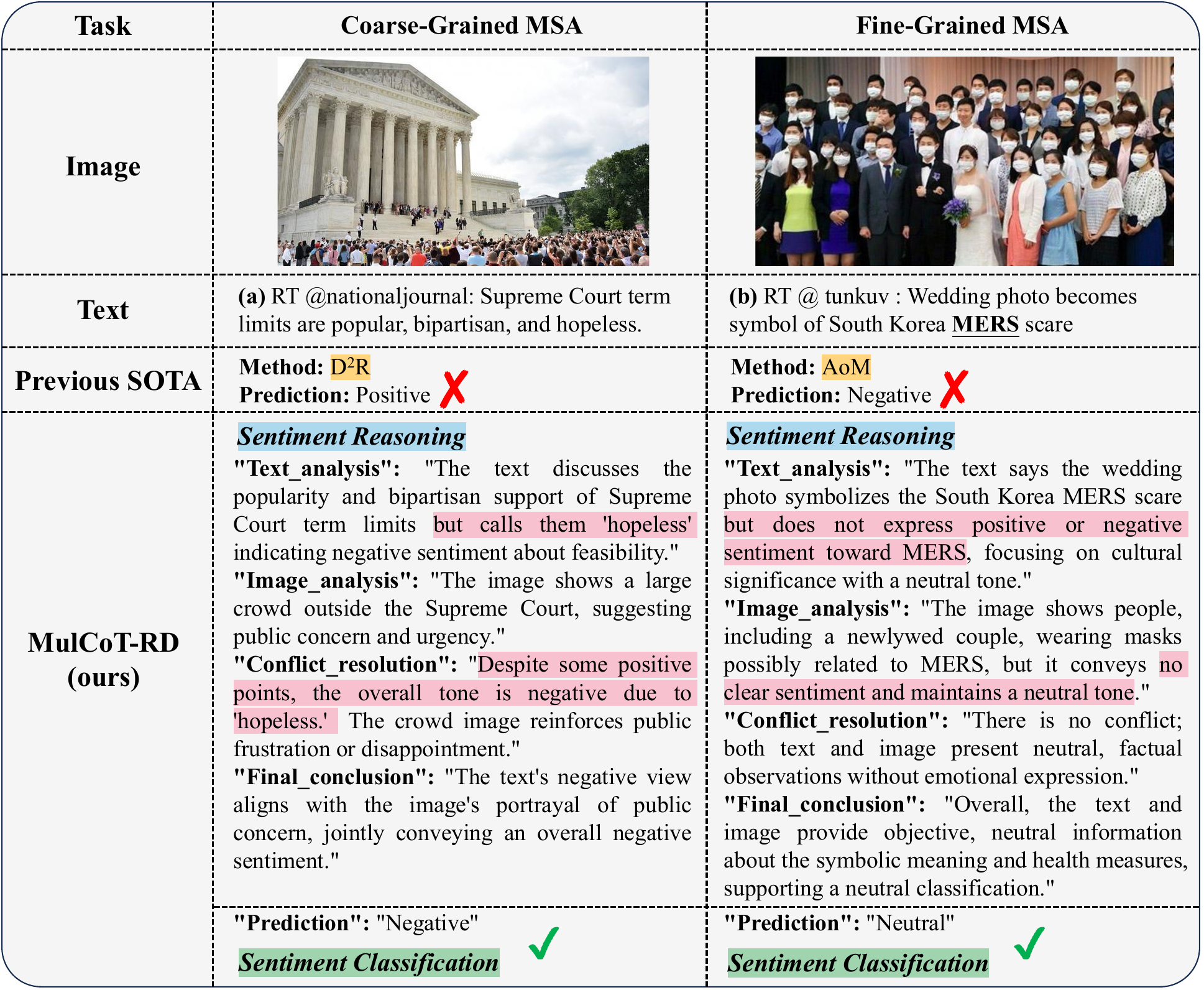}
\caption{Visualization of two samples, using the MulCoT-RD architecture with ID 4 from Table~\ref{table_four_architectures}.}
\label{figure_case_study}
\end{figure*}

To validate MulCoT-RD's effectiveness, we present two illustrative cases in Figure \ref{figure_case_study}. In case (a), $\mathbf{D}^2 \mathbf{R}$ incorrectly predicts sentiment by overrelying on surface-level positive terms like "popular" and "bipartisan" while missing the emotional shift from the word "hopeless" which establishes a negative tone. MulCoT-RD successfully captures this reversal. In case (b), the AoM misclassifies sentiment for the aspect term "MERS" by focusing on superficially negative words like "scare", leading to misinterpretation. MulCoT-RD effectively distinguishes between author stance (factual reporting) and content sentiment, producing correct predictions. This superior performance stems from our multi-task learning mechanism that integrates CoT reasoning and sentiment classification, enabling comprehensive modeling of intra-modal and cross-modal sentiment reasoning.

\section{Conclusion}

We focus on Joint Multimodal Sentiment Reasoning and Classification, JMSRC, in the resource-limited scenario that simultaneously generates multimodal reasoning chains and sentiment predictions. To address the dual challenges of reasoning interpretability and efficient deployment, we introduce MulCoT-RD, a unified framework combining structured CoT enhancement with reasoning distillation. Through a hierarchical teacher-assistant-student paradigm and joint multi-task learning, our method enables lightweight models to autonomously perform high-quality sentiment reasoning and classification. Extensive experiments across four datasets demonstrate the effectiveness and robustness of MulCoT-RD. In future work, we plan to incorporate direct preference optimization (DPO) with high- and low-quality reasoning sample filtering to further enhance the model's emotional reasoning quality and classification performance.

\section*{Limitations}

MulCoT-RD demonstrates strong performance on joint multimodal sentiment reasoning and classification with a lightweight model, yet several limitations exist. Our approach depends on high-performance teacher models (e.g., GPT-4o-mini) to synthesize the initial reasoning dataset via carefully crafted Chain-of-Thought prompts. This generation process entails considerable inference costs and is sensitive to the consistency and quality of the teacher model's outputs. The evaluation of generated reasoning chains relies mainly on automatic metrics, including n-gram overlap measures and embedding-based cosine similarity. These metrics may fail to adequately assess semantic coherence, factual accuracy, logical consistency, or human-perceived interpretability. Due to resource constraints, no large-scale human evaluation of reasoning quality was conducted. Although the student model is lightweight and targeted at resource-constrained deployment, it is built on a vision-language architecture that requires multimodal processing capabilities. This results in higher memory and computational demands than unimodal text-only models, which may limit deployment on extremely low-resource devices with minimal hardware acceleration support. We leave the exploration of more cost-effective reasoning data generation, comprehensive human evaluation of reasoning quality, multilingual generalization, and deployment on edge devices for future work.

\section*{Ethical considerations}

This work constructs a multimodal sentiment reasoning dataset by augmenting existing publicly available multimodal sentiment analysis datasets with model-generated reasoning produced by closed-source large multimodal language models (MLLMs). All source datasets are publicly available and are used in accordance with their original licenses and terms of use. The closed-source MLLMs are accessed exclusively through official APIs, and no private, proprietary, or user-identifiable data are accessed or collected.

The proposed dataset does not introduce new personal information beyond what is already present in the original public datasets, nor does it attempt to infer sensitive personal attributes. We acknowledge that sentiment interpretation is inherently subjective and that model-generated reasoning may reflect biases present in the underlying models. To reduce this risk, we employ structured prompting and verification procedures to improve annotation consistency and reliability.

The dataset is intended solely for research purposes in multimodal sentiment understanding and reasoning. We believe that this work adheres to established ethical standards for data usage and responsible application of large-scale models, and we do not anticipate foreseeable misuse beyond the scope of existing sentiment analysis research.

\section*{Acknowledgements}

The work is supported by the Fundamental Research Funds for the Central Universities under Grants (N25XQD004, N25ZLL045), and  the National Natural Science Foundation of China (Nos. 62272092, 62172086, 62576085). 

\bibliography{references}

\appendix

\section{Data expansion with Assistant Model}
\label{appendix_data_expansion}

After training the assistant model, we apply it to perform inference on the \textbf{original training set only}, explicitly excluding the validation and test sets to prevent any risk of label leakage. During this process, we retain only those samples whose predicted sentiment labels match the ground truth. These correctly predicted samples are then merged with the original training set to construct an expanded dataset, which is subsequently used for training the student model. Detailed results of the data expansion are presented in Table~\ref{table_results_assistant}.

\begin{table}[h]
\centering
\small
\setlength{\tabcolsep}{2.5pt}
\begin{tabular}{c| c| c c c c c c}
\toprule
\multirow{2}{*}{\textbf{Dataset}} & \multirow{2}{*}{\textbf{Samples}} & \multicolumn{3}{c}{\textbf{GPT-4o-mini}} & \multicolumn{3}{c}{\textbf{Qwen2.5-VL-72B}} \\
 & &  \textbf{Acc} & \textbf{w-F1} & \textbf{m-F1} & \textbf{Acc} & \textbf{w-F1} & \textbf{m-F1} \\
\midrule
MVSA-S & 3608 & 79.7 & 79.7 & 69.9 & 76.0 & 77.1 & 66.9 \\
MVSA-M & 13619 & 72.0 & 68.0 & 55.3 & 74.0 & 70.5 & 60.6 \\
Twitter-15 & 3179 & 94.0 & 94.1 & 92.6 & 95.6 & 95.6 & 94.6 \\
Twitter-17 & 3562 & 86.8 & 86.7 & 86.4 & 92.9 & 92.9 & 93.4 \\
\bottomrule
\end{tabular}
\caption{Performance of the Assistant Model (Qwen-2.5-VL-7B) on Training Sets During Data Expansion, Guided by Different Teacher Models.}
\label{table_results_assistant}
\end{table}

This strategy significantly increases the scale and diversity of the training data, broadens the coverage of sentiment label distributions, and incurs no additional manual annotation cost. It equips the student model with richer and higher-quality learning signals, effectively mitigating the challenge of limited annotated data commonly encountered in multimodal sentiment analysis tasks.

\section{Model Selection}
\label{appendix_model_selection}

To construct a hierarchical reasoning distillation framework for achieving efficient joint multimodal sentiment reasoning and classification (JMSRC), we carefully select the following models as the teacher model, the assistant model, and the student model. Table~\ref{table_model_selection} shows the specific model selections and their characteristics.
\begin{table}[h]
\centering
\small
\setlength{\tabcolsep}{5.5pt}
\begin{tabular}{c c c c c}
\toprule[1pt]
\textbf{Role} & \textbf{Model} & \textbf{Access} & \textbf{Release Date} \\
\midrule
\multirow{2}{*}{\textbf{Teacher}} & \cellcolor{gray!20}GPT-4o-mini & Closed & 2024.07 \\
& \cellcolor{gray!40}Qwen2.5-VL-72B    & Open    & 2025.02 \\
\multirow{2}{*}{\textbf{Assistant}} & Qwen3-VL-8B & Open & 2025.10 \\
 & Qwen2.5-VL-7B & Open & 2025.02 \\
\multirow{2}{*}{\textbf{Student}}   & Qwen3-VL-2B & Open & 2025.10 \\
   & Qwen2.5-VL-3B & Open & 2025.02 \\
\bottomrule
\end{tabular}
\caption{Model Selection and Characteristics. }
\label{table_model_selection}
\end{table}

\section{Baselines}
\label{appendix_baselines}

\textbf{Methods for coarse-grained MSA.} 1) \textbf{MultiSentiNet} \cite{xu2017multisentinet} is a deep attention-based semantic network for multimodal sentiment analysis. 2) \textbf{HSAN} \cite{xu2017analyzing} is a hierarchical semantic attentional network based on image captions for multimodal sentiment analysis. 3) \textbf{CoMN-Hop6} \cite{xu2018co} utilizes co-memory network to iteratively model the interactions between multiple modalities. 4) \textbf{MGNNS} \cite{yang2021multimodal} adopts multi-channel graph neural networks with sentiment-awareness for image-text sentiment detection. 5) \textbf{CLMLF} \cite{li2022clmlf} proposes a contrastive learning and multi-layer fusion method for multimodal sentiment detection. 6) \textbf{MVCN} \cite{wei2023tackling} designs a multi-view calibration network to solve the modality heterogeneity for multimodal sentiment detection. 7) \textbf{$\mathbf{D}^2 \mathbf{R}$} \cite{chen2024d2r} proposes a dual-branch dynamic routing network to enhance multimodal sentiment detection by effectively modeling cross-modal interactions. 8) \textbf{Emotion-LLaMA} \cite{cheng2024emotion} employs a specialized emotion tokenizer and instruction fine-tuning based on the LLaMA2-7B-chat to enhance multimodal emotion recognition. 9) \textbf{Qwen3-VL-8B-Thinking} \cite{yang2025qwen3} features Interleaved-MRoPE and DeepStack for powerful spatial-temporal reasoning, plus precise Text–Timestamp Alignment. With native 256K context, it excels at complex STEM and logic tasks.

\textbf{Methods for fine-grained MSA.} 1) \textbf{ESAFN} \cite{yu2019entity} is an entity-level sentiment analysis method based on LSTM. 2) \textbf{TomBERT} \cite{yu2019adapting} applies BERT to obtain aspect-sensitive textual representations. 3) \textbf{CapTrBERT} \cite{khan2021exploiting} translates images into text and construct an auxiliary sentence for fusion. 4) \textbf{JML} \cite{ju2021joint} is the first joint model for MABSA with an auxiliary cross-modal relation detection module. 5) \textbf{VLP-MABSA} \cite{ling2022vision} performs five task-specific pretraining tasks to model aspects, opinions, and alignments. 6) \textbf{CMMT} \cite{yang2022cross} implements a gate to control the multimodal information contributions during inter-modal interactions. 7) \textbf{AoM} \cite{zhou2023aom} introduces an aspect-oriented network designed to reduce visual and textual distractions from complex image-text interactions. 8) \textbf{Emotion-LLaMA} \cite{cheng2024emotion}. 9) \textbf{AETS} \cite{zhu2025aspect} improves multimodal sentiment analysis by enhancing aspects and simplifying text. 10) \textbf{Qwen3-VL-8B-Thinking} \cite{yang2025qwen3}

\section{Implementation Details}
\label{appendix_implementation_details}

\subsection{Hyperparameters in Multi-Task Learning}
\label{appendix_implementation_details1}

\begin{figure}[h]
\centering
\includegraphics[width=0.48\textwidth]{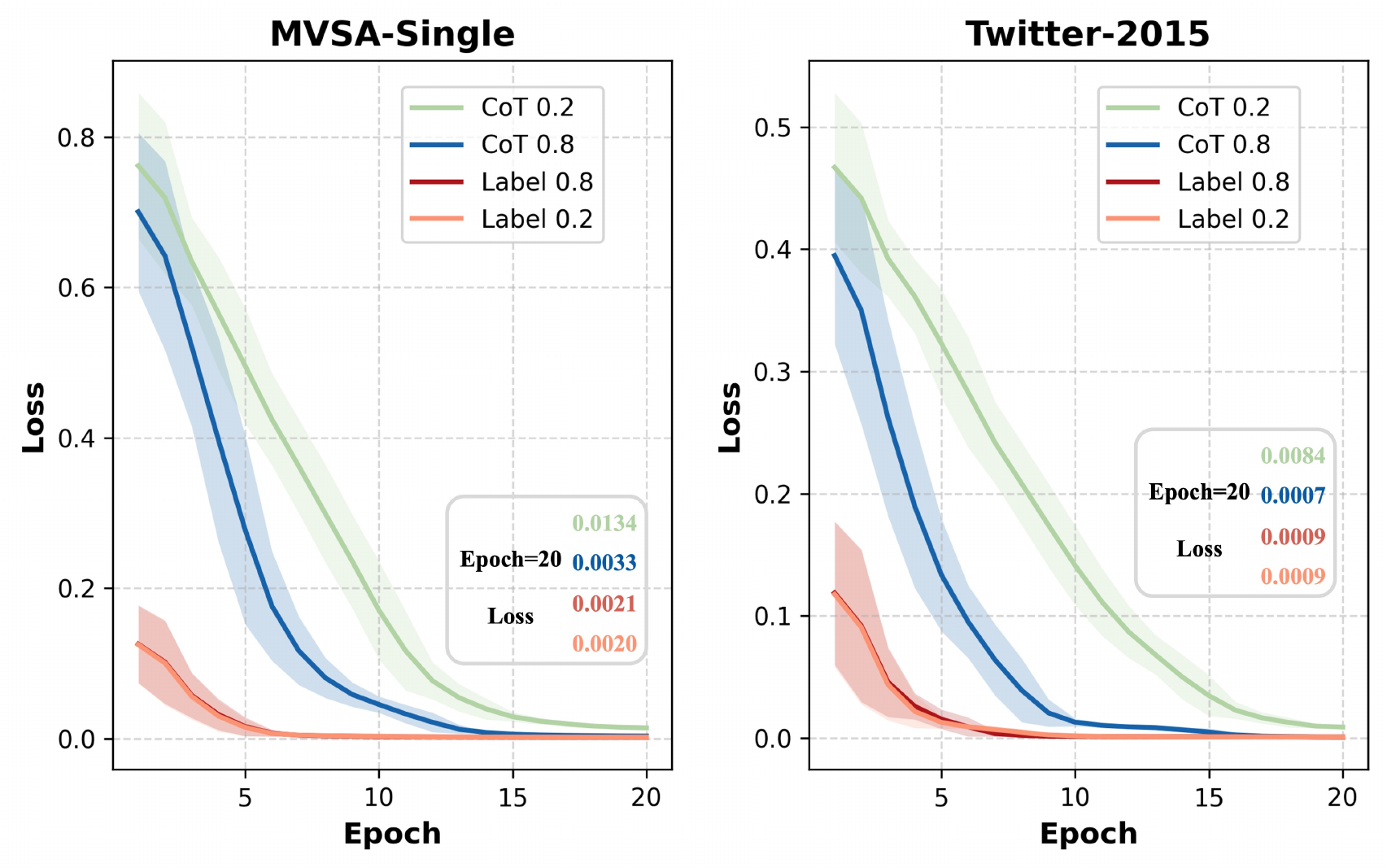} %
\caption{Effect of Loss Weight on Convergence for CoT Generation and Sentiment Classification Tasks.}
\label{figure_loss2}
\end{figure}

In our multi-task learning setup, we assign weights of 0.8 and 0.2 to the CoT (Chain-of-Thought) generation task and the sentiment classification task, respectively. This design is motivated by the following considerations:

\begin{itemize}
  \item \textbf{Task complexity}: CoT generation involves structured reasoning and belongs to a class of complex sequence generation tasks, which are more difficult to train and typically incur higher loss values. In contrast, sentiment classification is a relatively simple three-way classification task. Therefore, assigning a higher weight to CoT generation encourages the model to focus more on learning reasoning capabilities.
  
  \item \textbf{Convergence and gradient sensitivity}: As shown in Figure \ref{figure_loss2}, when the loss weights of the CoT generation task are set to 0.8 and 0.2, the model exhibits a significant difference in converged loss: specifically, the loss differs by approximately 4x on the MVSA-Single dataset (0.0134 vs. 0.0033) and by approximately 12x on the Twitter-2015 dataset (0.0084 vs. 0.0007). In contrast, for the sentiment classification task, the model’s converged loss remains largely consistent under different loss weight settings. Preliminary experiments show that the CoT task converges more slowly and is more sensitive to gradient fluctuations. Increasing its loss weight helps amplify gradient signals and improves training stability and task performance.
  
  \item \textbf{Empirical validation}: We experimented with different weight configurations (e.g., \{0.5, 0.5\}, \{0.2, 0.8\}) and observed that assigning lower weights to the CoT task led to slower loss reduction and decreased classification accuracy. In contrast, the \{0.8, 0.2\} setting consistently yielded better performance on both the validation and test sets.
\end{itemize}

This weighting scheme also reflects the task balancing principle proposed by CoTBal \cite{dai2024cotbal}, which emphasizes that in multi-task scenarios, loss weights should be adaptively assigned based on task complexity and learning dynamics to enhance main-task optimization and overall model performance.

\subsection{Hyperparameter in Knowledge Distillation}
\label{appendix_implementation_details2}

We set the hyperparameter $\lambda$ to 0.3, following the empirical practices in prior work \cite{lee2024mentor}, which achieve a good balance between stable training and effective knowledge transfer from the teacher model.

\subsection{LoRA Configuration}
\label{appendix_implementation_details3}

In all experiments, we adopt Low-Rank Adaptation (LoRA) for parameter-efficient fine-tuning of the multimodal models, rather than updating all parameters. In our implementation, the LoRA rank is set to $r=16$ and the scaling factor is set to $\alpha=32$. This strategy allows the models to adapt effectively to downstream tasks under limited computational resources, while significantly reducing the number of trainable parameters. The proportion of trainable parameters for each model is summarized in Table~\ref{table_lora_params}, and further implementation details are available in the GitHub repository.

\begin{table}[h]
\centering
\small
\setlength{\tabcolsep}{1.5pt} %
\begin{tabular}{cccc}
\toprule
\textbf{Model} & \textbf{Total Parameters} & \textbf{Trainable} & \textbf{Ratio (\%)} \\
\midrule
Qwen3-VL-8B    & 8{,}810{,}770{,}672 & 43{,}646{,}976 & 0.4954 \\
Qwen3-VL-2B    & 2{,}144{,}964{,}608 & 17{,}432{,}576 & 0.8127 \\
Qwen2.5-VL-7B  & 8{,}339{,}756{,}032 & 47{,}589{,}376 & 0.5706 \\
Qwen2.5-VL-3B  & 3{,}791{,}775{,}744 & 37{,}152{,}768 & 0.9798 \\
\bottomrule
\end{tabular}
\caption{Total and trainable parameter counts under LoRA fine-tuning.}
\label{table_lora_params}
\end{table}

\section{Details of the Adaptive Retry Controller (ARC)}
\label{appendix_arc}

Since large multimodal models are still prone to hallucinations in open-ended generation—such as producing content that deviates from the task specification or violating the required output format—the predicted sentiment label cannot always be reliably extracted from a single response. To mitigate this issue, we introduce an Adaptive Retry Controller (ARC) in the inference stage. Whenever the initially generated response does not conform to the expected format or fails to yield a valid sentiment label, ARC automatically triggers a retry process, prompting the model to regenerate the response for the same input sample. This process is repeated until a valid output is obtained or a predefined maximum number of retries is reached.

In our implementation, the maximum number of retries is set to three. This choice strikes a balance between improving label extraction robustness and controlling computational overhead. Empirically, approximately 79\% of the samples obtain a valid prediction within one or two generations, while only a small fraction require the full retry budget. As a result, ARC substantially reduces cases of missing or incorrectly parsed predictions, leading to a more stable and reliable inference pipeline without incurring prohibitive additional cost.

\section{Two-stage reasoning prompt template}
\label{appendix_two_stage_template}

\begin{figure}[t]
\centering
\includegraphics[width=0.48\textwidth]{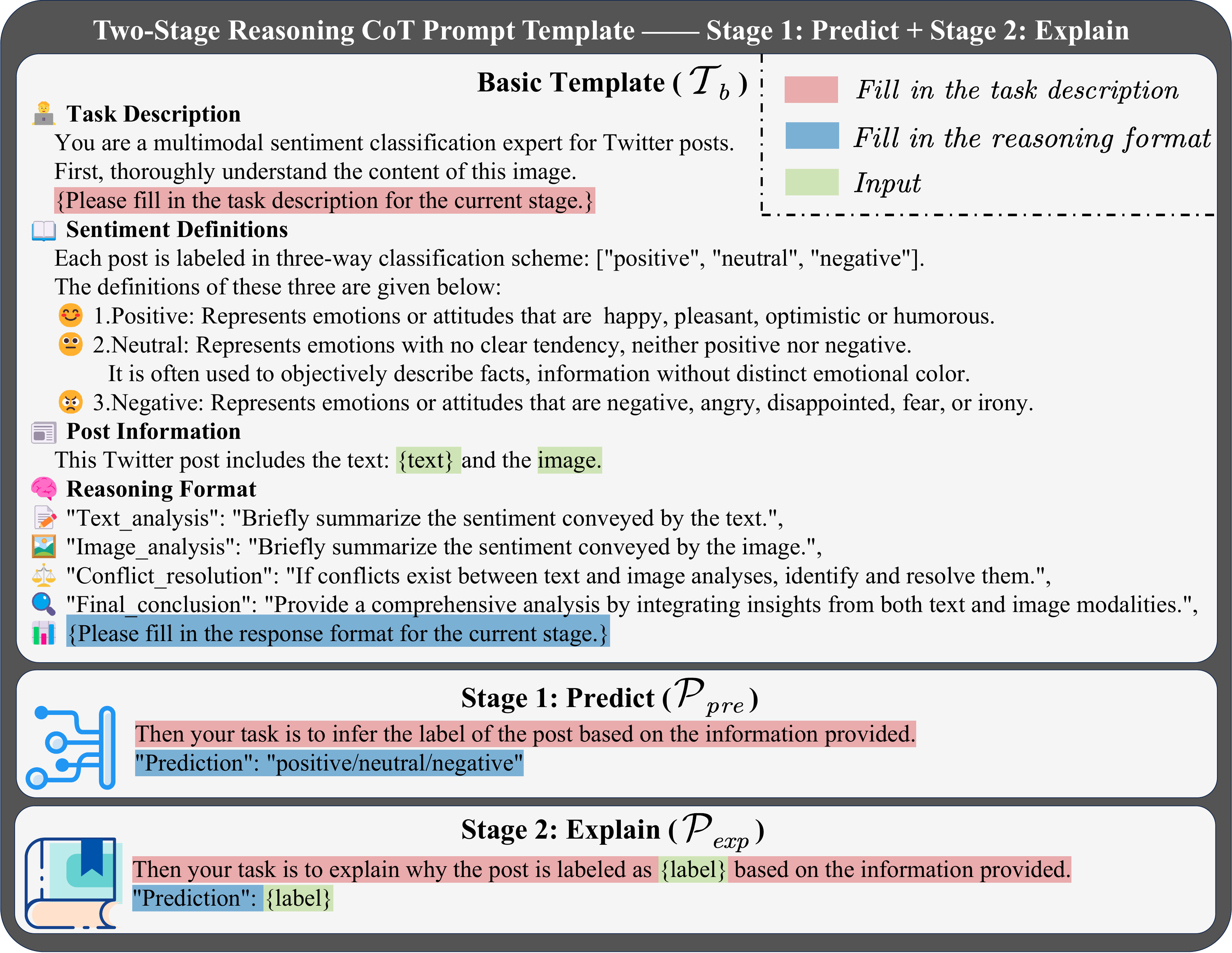}
\caption{Two-stage reasoning prompt template.}
\label{figure_two_stage_templates}
\end{figure}

The prompt template for the two-stage multimodal Chain-of-Thought (CoT) is illustrated in Figure \ref{figure_two_stage_templates}. By leveraging the basic template $\mathcal{T}_b$, the framework assigns a multimodal expert role to the model and prescribes a four-step structured reasoning process—comprising text analysis, image analysis, conflict resolution, and final conclusion—to explicitly address emotional correlations and conflicts between text and image modalities. Specifically, the execution is bifurcated into two stages: Prediction ($\mathcal{P}_{pre}$) and Explanation ($\mathcal{P}_{exp}$). In the first stage, the model is required to autonomously derive sentiment labels based on multimodal information to evaluate its inherent reasoning capabilities. In the second stage, the model is guided to backwardly construct logical justifications grounded in known labels, thereby generating deep-seated sentiment explanations. This two-stage decoupled design not only ensures the logical rigor of the reasoning chain but also substantially enhances the quality and interpretability of the generated data through the CoT mechanism.

\section{Robustness of MulCoT-RD}
\label{appendix_robustness}

\begin{figure}[h]
\centering
\includegraphics[width=0.48\textwidth]{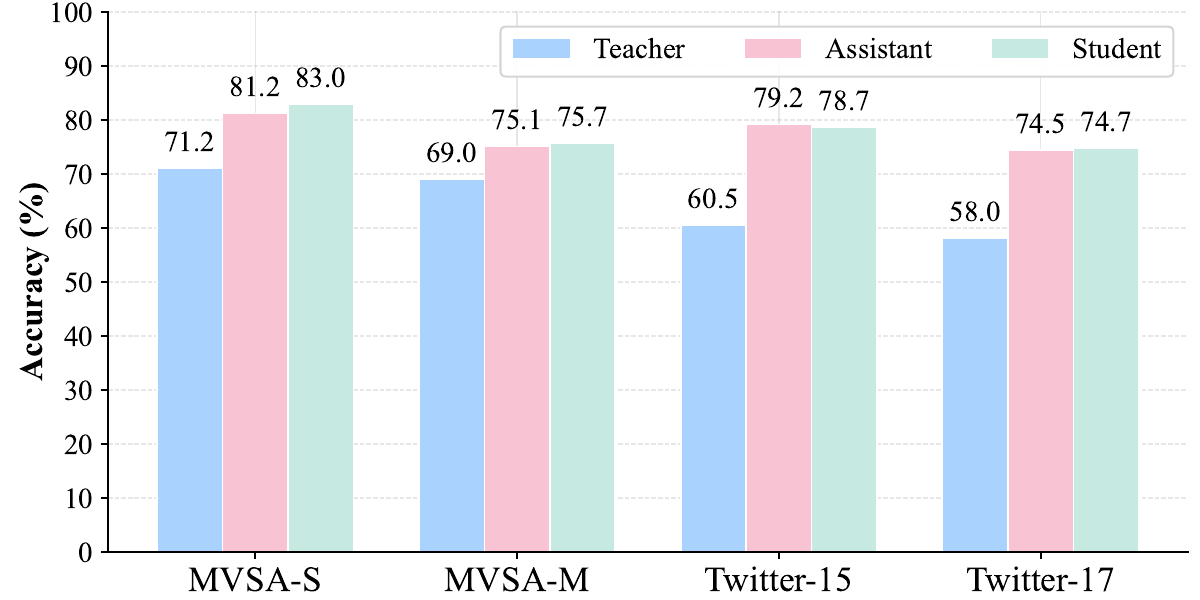}
\caption{Accuracy comparison of teacher (GPT-3.5-Turbo), assistant (Flan-T5-Large with 783M parameters) and student (Flan-T5-Base) models.}
\label{figure_flant5_acc}
\end{figure}

\begin{figure}[h]
\centering
\includegraphics[width=0.48\textwidth]{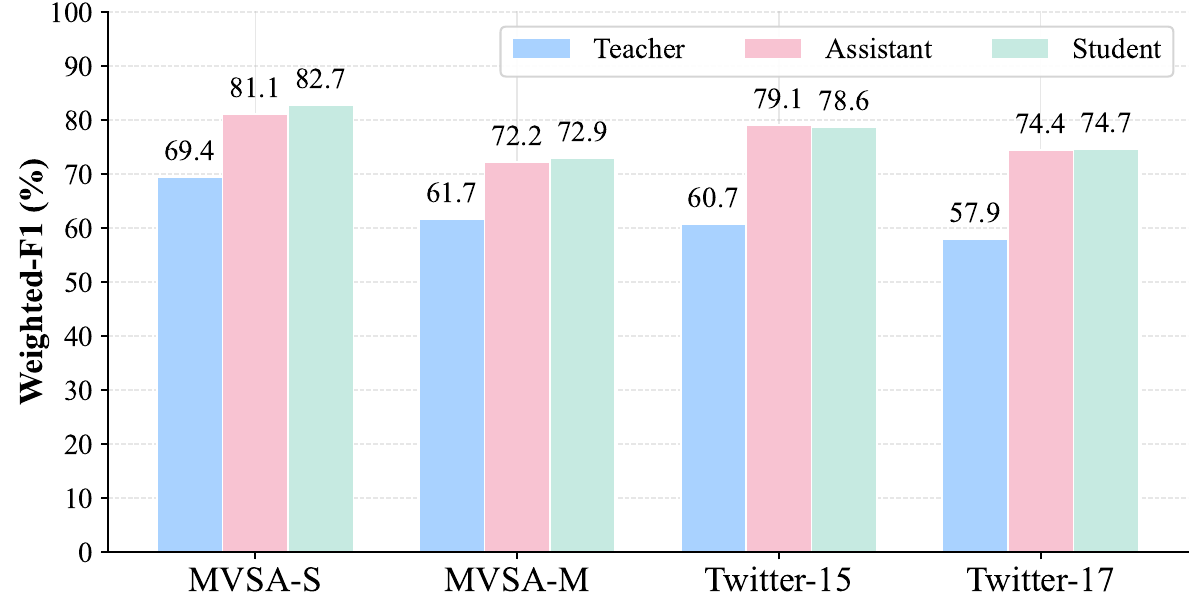} %
\caption{Weighted-F1 comparison of teacher(GPT-3.5-Turbo), assistant(Flan-T5-Large with 783M parameters) and student(Flan-T5-Base with 248M parameters) models.}
\label{figure_flant5_wf1}
\end{figure}

\begin{figure}[h]
\centering
\includegraphics[width=0.48\textwidth]{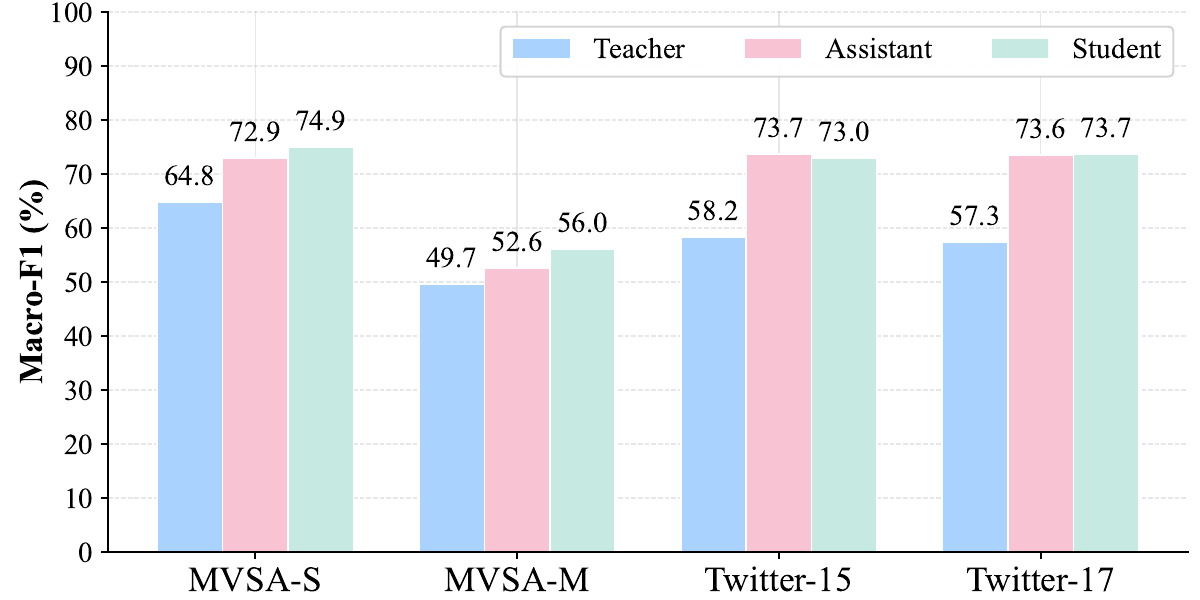} %
\caption{Macro-F1 comparison of teacher(GPT-3.5-Turbo), assistant(Flan-T5-Large with 783M parameters) and student(Flan-T5-Base with 248M parameters) models.}
\label{figure_flant5_mf1}
\end{figure}

For the Flan-T5 series. We utilize MiniCPM-o-2.6 \cite{team2025minicpm} to generate image captions, converting multimodal inputs to text-only format. Using the Flan-T5 architecture, we fine-tune both assistant and student models with full parameters, replicating the complete training pipeline including multimodal CoT enhancement, multi-task learning, and reasoning distillation. As shown in Figure~\ref{figure_flant5_acc} to \ref{figure_flant5_mf1}, the Flan-T5-based models achieve strong performance despite having only 248M parameters, demonstrating the robustness and adaptability of MulCoT-RD across diverse backbone architectures.

\section{Efficiency of MulCoT-RD}
\label{appendix_efficiency}

From the comparison of the data in Tables~\ref{table_efficiency_mvsa_single} and Table~\ref{table_efficiency_twitter_2015}, Qwen3-VL-2B demonstrates strong competitiveness under resource-constrained scenarios. On the MVSA-Single dataset, its GPU memory consumption is only 4.11 GB, approximately 13\% of Emotion-LLaMA (31.61 GB), substantially lowering the hardware deployment barrier. In contrast, our student models, by optimizing TFLOPs, achieve significantly reduced end-to-end inference latency while maintaining minimal memory usage. These results clearly indicate that knowledge distillation using high-quality two-stage CoT data enables lightweight models to achieve a favorable balance of high accuracy, low latency, and compact footprint in multimodal sentiment analysis tasks, providing reliable support for real-time online monitoring and mobile deployment.

\begin{table}[h]
\centering
\small
\setlength{\tabcolsep}{2.5pt}
\begin{tabular}{c|cccc}
\toprule
\multirow{2}{*}{\textbf{Model}} & \textbf{Params} & \textbf{TFLOPs} & \textbf{Latency} & \textbf{Memory} \\
                                & \textbf{(B)}      & \textbf{-}     & \textbf{(ms)}       & \textbf{(GB)} \\
\midrule
Qwen2.5-VL-72B & 73.41 & 38.24 & 17849.37 & 149.69 \\
Emotion-LLaMA & 14.58 & 4.11 & 11342.73 & 31.61 \\
Qwen3-VL-8B   & 8.81  & 3.13 & 10708.24 & 16.61 \\
Qwen3-VL-2B   & 2.14  & 3.15 & 8302.95  & 4.11  \\
Qwen2.5-VL-7B & 8.34  & 3.02 & 7002.99  & 15.81 \\
Qwen2.5-VL-3B & 3.79  & 1.32 & 8500.58  & 7.35  \\
\bottomrule
\end{tabular}
\caption{Efficiency comparison on the MVSA-Single dataset.}
\label{table_efficiency_mvsa_single}
\end{table}

\vspace{-0.5em}

\begin{table}[h]
\centering
\small
\setlength{\tabcolsep}{2.5pt}
\begin{tabular}{c|cccc}
\toprule
\multirow{2}{*}{\textbf{Model}} & \textbf{Params} & \textbf{TFLOPs} & \textbf{Latency} & \textbf{Memory} \\
                                & \textbf{(B)}      & \textbf{-}     & \textbf{(ms)}       & \textbf{(GB)} \\
\midrule
Qwen2.5-VL-72B & 73.41 & 38.41 & 17524.94 & 149.76 \\
Emotion-LLaMA & 14.58 & 4.31 & 11582.16 & 31.66 \\
Qwen3-VL-8B   & 8.81  & 3.39 & 9930.38 & 16.62 \\
Qwen3-VL-2B   & 2.14  & 3.60 & 8710.76  & 4.12  \\
Qwen2.5-VL-7B & 8.34  & 3.21 & 6209.65  & 15.83 \\
Qwen2.5-VL-3B & 3.79  & 1.44 & 8113.90  & 7.37  \\
\bottomrule
\end{tabular}
\caption{Efficiency comparison on the Twitter-2015 dataset.}
\label{table_efficiency_twitter_2015}
\end{table}

Furthermore, we observe that Qwen2.5-VL-7B exhibits even lower inference latency than Qwen3-VL-2B and Qwen2.5-VL-3B, despite having a larger parameter count. This can be attributed to two factors: first, Qwen2.5-VL-7B has fewer LLM layers (28 vs. 36) \cite{bai2025qwen2}; second, its larger hidden dimension increases the size of per-layer matrix operations, which allows modern GPUs to better saturate memory bandwidth and compute units, thereby achieving higher per-token computation efficiency (higher model bandwidth utilization) \cite{recasens2025mind}.

\end{document}